\documentclass[letterpaper]{article} 
\usepackage[]{AAAI2024/aaai24}  
\usepackage{times}  
\usepackage{helvet}  
\usepackage{courier}  
\usepackage[hyphens]{url}  
\usepackage{graphicx} 
\usepackage{natbib}  
\usepackage{caption} 
\usepackage{amsmath,amsfonts,bm}

\def\eqref#1{equation~\ref{#1}}

\def\1{\bm{1}}

\DeclareMathAlphabet{\mathsfit}{\encodingdefault}{\sfdefault}{m}{sl}
\SetMathAlphabet{\mathsfit}{bold}{\encodingdefault}{\sfdefault}{bx}{n}

\newcommand{\norm}[1]{\left\|#1\right\|}

\usepackage[vlined,ruled]{algorithm2e}

\usepackage[utf8]{inputenc} 
\usepackage[T1]{fontenc}    
\usepackage{booktabs}       
\usepackage{amsfonts}       
\usepackage{nicefrac}       
\usepackage{microtype}      

\usepackage{multirow}
\usepackage{hhline}
\usepackage{enumitem}
\usepackage{amsmath,mathtools}

\makeatletter
\newcommand{\raisemath}[1]{\mathpalette{\raisem@th{#1}}}
\newcommand{\raisem@th}[3]{\raisebox{#1}{$#2#3$}}
\makeatother

\title{Knowledge Propagation over Conditional Independence Graphs}
\author{
    Urszula Chajewska \&  
    Harsh Shrivastava
}
\affiliations{

    ~~Microsoft Research, Redmond, USA\\
    ~~\{urszc, hshrivastava\}@microsoft.com
}

\begin{document}

\maketitle




\newcommand{\fix}{\marginpar{FIX}}
\newcommand{\new}{\marginpar{NEW}}
\newcommand\independent{\protect\mathpalette{\protect\independenT}{\perp}}
\def\independenT#1#2{\mathrel{\rlap{$#1#2$}\mkern2mu{#1#2}}}

\newcommand{\Rho}{\mathrm{P}}


\begin{abstract}
Conditional Independence (CI) graph is a special type of a Probabilistic Graphical Model (PGM) where the feature connections are modeled using an undirected graph and the edge weights show the partial correlation strength between the features. Since the CI graphs capture direct dependence between features, they have been garnering increasing interest within the research community for gaining insights into the systems from various domains, in particular discovering the domain topology. In this work, we propose algorithms for performing knowledge propagation over the CI graphs. Our experiments demonstrate that our techniques improve upon the state-of-the-art on the publicly available Cora and PubMed datasets. 

\textit{Software}:~\url{https://github.com/Harshs27/reasoning-CI-graphs}
\end{abstract}

\section{Introduction}

Probabilistic Graphical Models (PGMs) are a useful tool for domain exploration and discovery of
domain structure. They rely on probabilistic independence and conditional independence assumptions
between features (nodes) to make representation, learning, and inference feasible even in domains
with a large number of features. Apart from computational benefits, such independence properties
give insight into relations between features in the domain. Conditional Independence graphs are a
type of PGMs that model direct dependencies between the input features as an undirected graph. Each edge represents partial correlations
between the connected node features (definition taken from a recent survey~\cite{shrivastava2022methods}). CI graphs are
primarily used to gain insights about the feature relationships to help with decision making. 
For
instance, CI graphs were used to analyze Gene Regulatory Networks~\cite{shrivastava2019glad}, detecting wind-speed in North America~\cite{greenewald2019tensor}, understanding brain
connectivity in context of identifying differences among autistic patients vs normal subjects~\cite{pu2021learning}, for increasing methane yield
in anaerobic digestion~\cite{shrivastava2022a}, 
analysing the infant-mortality data from CDC in the US~\cite{shrivastava2022neural,shrivastava2023neural} and many more. In some cases, they are
also used to study the evolving feature relationships with time~\cite{hallac2017network,imani2023uglad}.

Apart from feature values used to learn the domain structure and represent it as a graph, individual features may have a number of other attributes.  For example, we could use a paper's word count to discover a graph where each node would represent a paper and similar papers would be connected by an edge.  We can assume that a paper's word count data is predictive of the paper's subject matter.  If we know the subject of some fraction of the papers in a collection, we can use the graph to discover the subject of the remaining papers based on the graph's connection pattern.  This process is known as knowledge propagation (KP) in graphs.  

In this work, we focus on developing algorithms for doing knowledge propagation over Conditional Independence (CI) graphs. This is an important tool that is useful for analysing a wide variety of domains. We present multiple knowledge propagation algorithms that are either based on modifying the existing approaches or on utilizing the underlying probabilistic graphical model formulation of CI graphs to obtain an analytical solution. 




\section{Background and Related work}

\subsection{Obtaining Conditional Independence Graphs}

\begin{figure*}
\centering 
\includegraphics[width=135mm]{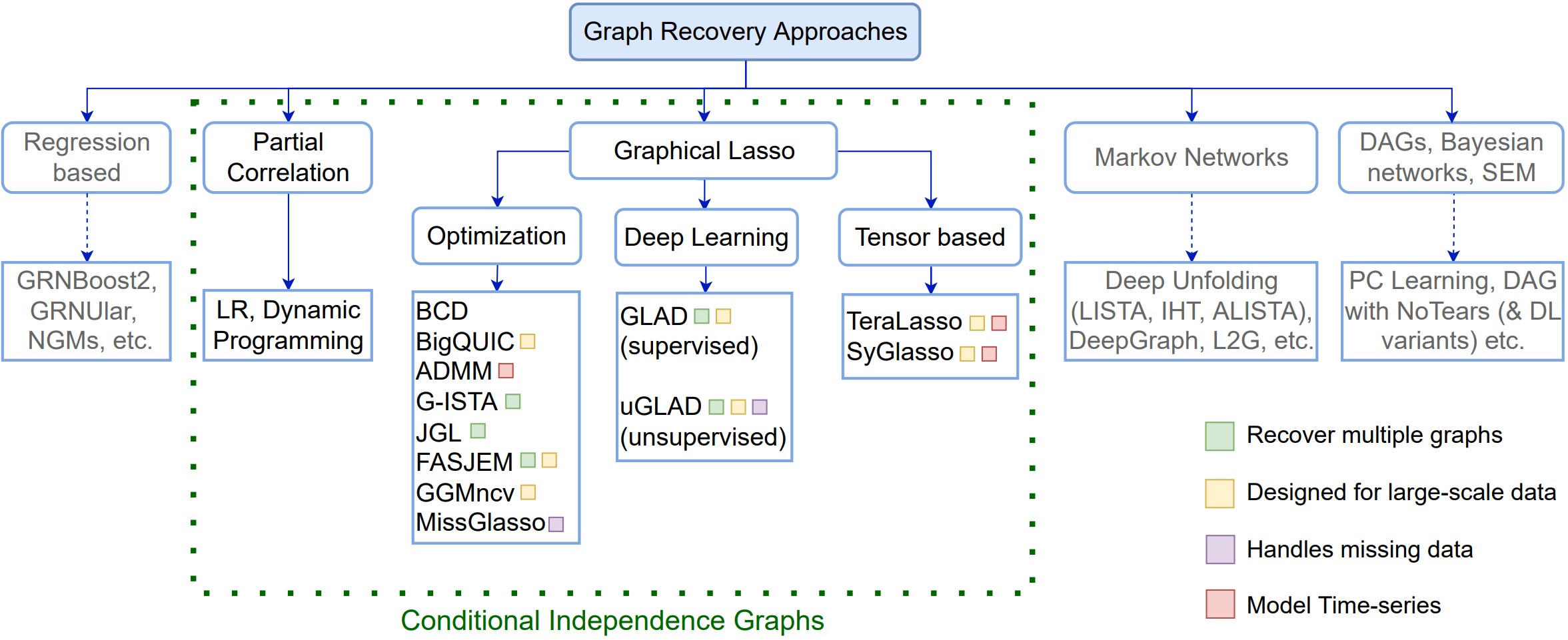}
\caption{\small \textbf{Graph Recovery approaches.} Methods used to recover Conditional Independence graphs, borrowed from~\cite{shrivastava2023methods}. Please refer to this survey for details. The algorithms (leaf nodes) listed here are representative of the sub-category and the list is not exhaustive.}
\label{fig:ci-graph-diagram}
\end{figure*}

Mathematically, the CI graph of a set of random variables $X_i$ 
is the undirected graph $G=(D,E)$ where $D=\{1,2,…,d\}$ is the set of nodes (variables) and $(i,j)$ is not in the edge set if and only if $X_i\independent X_j|\textbf{X}_{D\backslash{i,j}}$, where $\textbf{X}_{D\backslash{i,j}}$ denotes the set of all of the random variables except for $X_i$ and $X_j$. The edges of the CI graph represent the partial correlations between the features (nodes) and their magnitude ranges between 
$(-1,1)$, refer to~\cite{shrivastava2023methods}. 

Recovery of conditional independence graphs is a topic of wide research interest. Few prominent formulations to recover CI graphs include modeling using (1) Linear regression, (2) Recursive formulation, and (3) Matrix inversion approaches. Based on the problem formulation, many different optimization algorithms have been developed with each having their own capabilities and limitations. Fig.~\ref{fig:ci-graph-diagram} summarizes the existing graph recovery approaches. Besides, tools like a differentiable \textit{F-beta} score~\cite{chen2020rna,shrivastava2020grnular,shrivastava2022grnular}, clubbed with loss-balancing technique~\cite{rajbhandari2019antman} can be useful while training CI graph recovery models to obtain desired fit to the data while maintaining the desired sparsity. Technique proposed in~\cite{aluru2021engrain} can be extended to obtain a consensus from the ensemble of multiple CI graphs. Such requirements can arise in understanding gene regulatory networks or time-series data like stock markets or dynamic time-varying systems. 

\subsection{Knowledge Propagation in Graphs}

The problem of knowledge propagation (KP), also known as attribute propagation or label propagation, has been well studied in the past years. Recent survey by~\cite{garza2019community} discusses numerous techniques developed for the community detection problem.~\cite{iscen2019label} proposed a method to do label propagation for deep semi-supervised learning.~\cite{fujiwara2014efficient} proposed ways to do efficient label propagation over large-scale graphs. Works done by ~\cite{zhu2005semi,gregory2010finding,gong2016label} exploit the different levels of label importance in a particular domain to design their propagation algorithms. Researchers have used KP techniques for various interesting applications ranging from community detection to understanding dynamics of anaerobic digestion, finance \& healthcare problems. A common requirement for the KP algorithms for tackling problems in these domains is an existence of an underlying graph, which is either provided by an expert or algorithmically discovered. As the sparse graph recovery techniques are advancing~\cite{shrivastava2019glad,pu2021learning,shrivastava2022a}, KP algorithms have been explored for various newer domains. 

Some algorithms developed for knowledge propagation are designed specifically for knowledge graphs.  We do not review them here, as it is out of scope for this work. Of particular interest are Graph Neural Network (GNN) methods, which can be applied to a wider range of graphs and produce state-of-the-art results. GNN-based methods learn a mapping of nodes to a low dimensional space that maximizes the likelihood of preserving graph neighborhood~\cite{Perozzi14deepwalk,Tang15line,grover2016node2vec}. These node representations are used to predict node attribute values. The algorithms differ in graph properties taken into account and neighborhood definitions, with~\cite{grover2016node2vec} being the most flexible in this regard.

Recent developments in video representation and video generation domains have increasingly seen a surge of graph based approaches, especially probabilistic graphical models and fitting temporal processes~\cite{oprea2020review}. Videos can be interpreted as a sequential collection of frames and graph based methods can potentially model the feature interactions within a frame. Tracking these changes across frames is translated to the corresponding dynamic changes in the feature graphs~\cite{shrivastava2024methods}. Many different video tasks like action recognition, prediction of interaction activity, video segmentation \& object detection~\cite{bodla2021hierarchical,jiao2021new,zhou2022survey,saini2022recognizing} can benefit from knowledge propagation over the graphs to gain enhanced insights. Fitting temporal processes and conditional independence graphs to model the inter-frame relations can especially benefit from efficient knowledge propagation techniques. Examples include video prediction by continuous multi-dimensional processes~\cite{shrivastava2024video1}, video priors representation~\cite{shrivastava2024video2,shrivastava2024video3} and generating diverse video frames~\cite{denton2018stochastic,shrivastava2021diverse,shrivastava2021diversethesis}. 

\subsection{Preprocessing the data}

We briefly cover some essential preliminary procedures needed to make sure that the CI graph recovery algorithms can discover meaningful relationships.  

\textbf{Normalization.} It is often desirable to normalize the input before feeding it to the recovery algorithms. This is primarily needed for the stability of the optimization. Some of the common normalization procedures include \textit{min-max}, \textit{mean}, \textit{centered log-ratio}, \textit{additive log-ratio}, \textit{standardized moment}, \textit{student's t-statistic} etc.

\textbf{Handling mixed input datatypes.} Encountering categorical variables along with numerical ones in the input is very common. In such cases, calculating the covariance matrix (used as an input to many CI recovery approaches) between the features can be tricky. The covariance matrix calculations can be categorized into calculating the following types of correlations or associations: (I) numerical-numerical correlation can be captured using Pearson's correlation coefficient. Depending on the type of non-linearity desired, one can also opt for Spearman Correlation, Kendell's Tau, Somer's D among others. (II) Categorical-categorical association can be calculated by using Cram\'ers V statistic including the bias correction. (III) Numerical-Categorical correlations are often evaluated using correlation ratio, point biserial correlation or Kruskal-Wallis test by ranks also known as H test.



\section{Knowledge Propagation over CI graphs}




\begin{algorithm}
  \DontPrintSemicolon
  \SetKwProg{Fn}{Function}{:}{}
  \SetKwFor{uWhile}{While}{do}{}
  \SetKwFor{ForPar}{For all}{do in parallel}{}
  \SetKwFunction{iterativeexp}{iterative-exp}
  \SetKwFunction{iterativeposneg}{iterative-posneg}
  \SetKwFunction{analytical}{analytical}
  \SetKwFunction{propagate}{KnowProp}
    \Fn{\iterativeexp{$\Rho^e, \textbf{n}^0$}}{
        $\textbf{n}_K+\textbf{n}_U^0 \gets \textbf{n}^0$, split the data \;
        $\Rho_U^e\gets\Rho^e[\textbf{n}_U^0] \in \mathbb{R}^{U\times D}$, unknown\;
        $t=0$\;
      \uWhile{$\norm{\textbf{n}^t-\textbf{n}^{t-1}}_2^2>\epsilon$}{
            $\textbf{n}^{t}_U\leftarrow\langle\Rho^e_U, \textbf{n}^{t-1}\rangle$\;
            $\textbf{n}^t  \gets \textbf{n}_K+\textbf{n}_U^t $\; 
            $t = t+1$\;
        }
     \KwRet $\textbf{n}^t$
    }\vspace{2mm}
    \Fn{\iterativeposneg{$\Rho^+, \Rho^-, \textbf{n}^0$}}{
        $\textbf{n}_K+\textbf{n}_U^0 \gets \textbf{n}^0$, split the data \;
        $\Rho_U^+\gets\Rho^+[\textbf{n}_U^0] \in \mathbb{R}^{U\times D}$, unknown\;
        $\Rho_U^-\gets\Rho^-[\textbf{n}_U^0] \in \mathbb{R}^{U\times D}$, unknown\;
        $t=0$\;
      \uWhile{$\norm{\textbf{n}^t-\textbf{n}^{t-1}}_2^2>\epsilon$}{
            $\textbf{n}^{t}_U\leftarrow\langle\Rho^+_U, \textbf{n}^{t-1}\rangle$\;
            \quad+ $\texttt{KL}(\textbf{n}^0_U,\langle\Rho^+_U, \textbf{n}^{t-1}\rangle)$\;
            \quad+ $\texttt{KL}(\textbf{n}^0_U,\langle\Rho^-_U, \textbf{n}^{t-1}\rangle)$\;
            $\textbf{n}^t  \gets \textbf{n}_K+\textbf{n}_U^t$\; 
            $t = t+1$\;
        }
     \KwRet $\textbf{n}^t$
    }\vspace{2mm}
    \Fn{\analytical{$\Rho^e, \textbf{n}^0$}}{
        $\textbf{n}_K+\textbf{n}_U^0 \gets \textbf{n}^0$, split the data \;
        $\left[\Theta_{UU}^e, \Theta_{UK}^e\right]\gets\Rho^e$\;
        $\textbf{n}_U = \left(I-\Rho^e_{UU}\right)^{-1}\Rho^e_{UK}\cdot \textbf{n}_K$\;
        $\textbf{n}\gets \textbf{n}_K+\textbf{n}_U$\;
     \KwRet $\textbf{n}$
    }\vspace{2mm}
    \Fn{\propagate{$\Rho^e, \Rho^+, \Rho^-, C$}}{
        Input: $\Rho^e, \Rho^+,\Rho^-\in\mathbb{R}^{D\times D}$, $C\in\operatorname{cat}^{D\times 1}$\;
        $\{K, U\} \gets C$,  {\small split into [known, unknown]}\;
        $\textbf{n}_K \gets \delta_{i,K}$ (delta dist.)\;
        $\textbf{n}_U \gets \mathcal{U}(U)$ (uniform dist.)\;
        $\textbf{n}^0 \gets \textbf{n}_K + \textbf{n}_U \in\mathbb{R}^{D\times C}$\;
        $\textbf{n} \gets $\iterativeexp($\Rho^e, \textbf{n}^0$)\;
        $\quad \cdots $ or $\cdots$\;
        $\textbf{n} \gets $\iterativeposneg($\Rho^+,\Rho^-,\textbf{n}^0$)\;
        $\quad \cdots $ or $\cdots$\;
        $\textbf{n} \gets $\analytical($\Rho^e, \textbf{n}^0$)]\;
        \KwRet  $\textbf{n}$
    }
\caption{KP over a CI graph}\label{algo:prop-algo}
\end{algorithm}

Given a CI graph and feature attributes for some of the nodes, we want to propagate information among the graph nodes to predict the attribute values for the remaining nodes. 





Note that knowledge propagation can be used for input data features or other node properties highly correlated with input data.  For example, consider an anaerobic digestion domain with input data listing organisms' abundance in various samples.  In the CI graph built for this domain, correlations between nodes (organisms) indicate which microbes tend to grow and diminish in number in similar environments.  We can use knowledge propagation to predict missing organisms' abundance for samples where only some of the organisms are listed.  We can also use it to predict correlated features of organisms, such as microbial function, on the assumption that organisms that thrive in similar environments share this property. Interestingly knowledge propagation over CI graphs can be leveraged to improved interpretability and performance of text mining tools~\cite{roche2017valorcarn,fize2017geodict,antons2020application}.

Throughout our analysis, we assume access to a partial correlation matrix $\Rho\in\mathbb{R}^{D\times D}$ obtained after running a CI graph recovery algorithm over the input data $X$ with $M$ samples and $D$ features. For ease of notation, we abbreviate $\Rho_{i,j} = \rho_{X_i, X_j}.\textbf{X}_{D\backslash{i,j}}$ in our derivations (refer to Sec.2.2 of ~\cite{shrivastava2023methods} for a complete  mathematical treatment). 
We use $\textbf{n}\in\mathbb{R}^{D\times C}$ to denote a distribution over all possible $C$ categories of an attribute for each of $D$ nodes. 
The nodes can be divided into known and unknown sets, $\textbf{n} = \textbf{n}_K+\textbf{n}_U$, where $\textbf{n}_K$ correspond to nodes for which the categories are known and $\textbf{n}_U$ to unknown nodes. If the category is known with certainty, then the corresponding $n_i$ entry is a delta distribution $\delta_{i, c}$ where the entry is $1$ for the known category and $0$'s for the rest. Users can also specify a distribution over the categories of an attribute for known nodes based on their prior knowledge. The KP task is to predict the most likely categories for the unknown nodes.
\subsection{Transition probability matrix}
As a first step, we convert the partial correlation matrix into a transition probability matrix.

One approach to creating a transition probablity matrix is to take entry-wise exponential and then perform a row-wise normalization:
\begin{align}\label{eq:normalized-exp-theta}
    \Rho^e_{i,j} = \frac{e^{\alpha\Rho_{i,j}}}{\sum_{j}^De^{\alpha\Rho_{i,j}}}
\end{align}
where $\alpha$ is the scaling intensity parameter. This particular transformation to $\Rho^e$ 
exponentially scales down the weights of the negative correlations turning them into less favorable paths to be traversed for knowledge propagation which is the desired behaviour.

Note that the transformation above causes edges with negative correlation values to influence the outcome in a positive way, albeit much more weakly than edges with positive correlation values. It should work well in domains where the attribute correlations strongly mimic node value correlations.  However, in other domains, we might use an alternative conversion.  We start by splitting the partial correlation matrix into matrices with only positive and only negative correlations:
\begin{align}\label{eq:pos}
    \Rho^+_{i,j} =
    \begin{cases}
    \Rho_{i,j} & \Rho_{i,j} > 0 \\
    0 & \text{otherwise.}
\end{cases}
\end{align}
and
\begin{align}
    \Rho^-_{i,j} =
    \begin{cases}
    -1*\Rho_{i,j} & \Rho_{i,j} < 0 \\
    0 & \text{otherwise.}
\end{cases}
\end{align}
$\Rho^{+}$ and $\Rho^{-}$ are then normalized row-wise:
\begin{align}\label{eq:normalized-pos-neg}
    \Rho^+_{i,j} = \frac{\Rho^+_{i,j}}{\sum_{j}^D{\Rho^+_{i,j}}}, \qquad \Rho^-_{i,j} = \frac{\Rho^-_{i,j}}{\sum_{j}^D{\Rho^-_{i,j}}}
\end{align}




\subsection{Iterative approach} 
We use either transition probability matrix $\Rho^e$ from Eq.~\ref{eq:normalized-exp-theta} or the two positive and negative transition probability matrices from Eq.~\ref{eq:normalized-pos-neg} to propagate known labels among the unknown features by a diffusion process where the labels of the known features are fixed, following Alg.~\ref{algo:prop-algo}. The propagation has a generic update of the form of $\textbf{n}^{t}_U\leftarrow\Rho^e\cdot\textbf{n}^{t-1}$ for $\Rho^e$.  In case of positive and negative transition matrices, we can choose among a range of more complex updates. The components include updates based on the two transition matrices:  $\textbf{n+}^t_U \leftarrow \Rho^+\cdot\textbf{n}^{t-1}$ and $\textbf{n-}^t_U \leftarrow \Rho^-\cdot\textbf{n}^{t-1}$ and two regularization terms based on Kullback-Leibler divergence (or another distribution distance such as Wasserstein distance) for these two with respect to the uniform distribution or $\textbf{n}^{t-1}_U$. An update form that worked particularly well in our experiments is:
\begin{align} \label{eq:iterative-update}
\textbf{n}^{t}_U \leftarrow & \Rho^+\cdot\textbf{n}^{t-1} + D_{KL}(\textbf{n}^0_U, \Rho^+\cdot\textbf{n}^{t-1}) \nonumber \\
& + D_{KL}(\textbf{n}^0_U, \Rho^-\cdot\textbf{n}^{t-1})
\end{align}
where $D_{KL}$ is Kullback-Leibler divergence and $\textbf{n}^0_U$ holds the uniform distribution initialized at the beginning of the algorithm.  The inclusion of distance from the uniform distribution in the update based on the positive matrix serves to slow the update speed; distance from uniform of the update based on the negative matrix takes into account the influence of negative partial correlations.
We run this algorithm until convergence or for a fixed number of iterations.
After the convergence of the algorithm at iteration $T$, we can apply a row-wise softmax operation to get the final predicted categories as
    $\textbf{n}_U^c = \operatorname{row-argmax}\left(\operatorname{softmax}(\textbf{n}^{T}_U)\right)$
. The function $\operatorname{argmax}$ defines a strategy to select the most appropriate category from the obtained converged row-wise distribution. Other selection strategies can also be tried out based on the results on the stand-out data. For example, we can only select the most likely category in cases where the probability for the category with the highest score exceeds a given threshold.

\subsection{Analytical solution} \label{sec:analytical-proof}

\textit{Objective}: Given an assignment of the attributes for some nodes and the partial correlation matrix of the underlying CI graph, our aim is to derive an analytical solution to predict the attribute distribution of the remaining nodes. 

\textit{Overview of approach}: The idea behind the proof is to simulate the iterative approach until convergence (or infinite number of times). The analytical solution derived below is theoretically equivalent to running the iterative method until convergence. We follow the proof of convergence procedure of a label propagation algorithm presented in the Chapter 2 of \cite{zhu2005semi}. Specifically, the novelty of our method lies in our procedure of obtaining the transition probability matrix from the precision matrix. We were able to show that the transition probability matrix satisfies the properties needed to successfully obtain an analytical solution. 

\textit{Claim}: The following update
\begin{align}
    n_U = \left(I-\Rho^e_{UU}\right)^{-1}\Rho^e_{UK}\cdot n_K
\end{align}
is equivalent to running the iterative update 
\begin{align}\label{eqn:itr-update}
    n_U^t\gets \Rho^e_{UU}\cdot n_U^{t-1} + \Rho^e_{UK} \cdot n_K
\end{align} 
infinite number of times.

\textit{Proof}: Starting with the partial correlation matrix $\Rho$, we apply entrywise exponential of matrix $e^{\alpha\Rho}$ and then perform row-wise normalization of the $e^{\alpha\Rho}$ matrix. 
 The resultant matrix obtained is our transition probability matrix $\Rho^e$, refer to Eq.~\ref{eq:normalized-exp-theta}. In this way, the negative correlation weights are scaled down considerably compared to the positive correlations and the weights are converted to positive values. This is the desired behaviour as we want the positive partial correlations to have more influence than no correlations and the no correlations to have more weight than the negative partial correlations. Note that due to exponentiation, correlation entries between nodes that are not connected in the original graph all have the value of 1 before normalization.


Let the nodes be divided into known and unknown categories as $n$=$\begin{bmatrix}
    n_K\\n_U
\end{bmatrix}$ and $\Rho^e$=$\begin{bmatrix}
    \Rho^e_{KK} & \Rho^e_{KU}\\
    \Rho^e_{UK} & \Rho^e_{UU}
\end{bmatrix}$


Continuing the update given in Eq.~\ref{eqn:itr-update} in the iterative procedure,  
will lead to 
\begin{align}
    n_U = \lim_{t\rightarrow \infty} \left(\Rho^e_{UU}\right)^t\cdot n^0_U + \left(\sum_{i=1}^t\left(\Rho_{UU}^e\right)^{(i-1)}\right)\Rho^e_{UK}\cdot n_K
\end{align}
We can show that the first term $\lim_{t\rightarrow \infty} \left(\Rho^e_{UU}\right)^t\cdot n^0_U \rightarrow 0 $. $\Rho_{UU}^e$ is a sub-matrix of a row-normalized $\Rho^e$ and  
we consider the row-wise sums of the sub-matrix
\begin{align}
    	\exists \mu < 1, \sum_{j=1}^D \left(\Rho^e_{UU}\right)_{i,j}\leq\mu, ~\forall{i}\in{1,\cdots,D}
\end{align}
Thus, 
\begin{align}
    \sum_{j}\left(\Rho^e_{UU}\right)^t_{i,j} &= \sum_j\sum_k\left(\Rho_{UU}^e\right)^{(t-1)}_{i,k}\left(\Rho^e_{UU}\right)_{k,j} \nonumber \\
    & = \sum_k\left(\Rho_{UU}^e\right)^{(t-1)}_{i,k}\sum_j\left(\Rho^e_{UU}\right)_{k,j}\\
    & \leq \sum_k\left(\Rho_{UU}^e\right)^{(t-1)}_{i,k} \mu
    \leq \mu^t\nonumber
\end{align}
This result shows us that the choice of the initial distribution over the categories $n^0_U$ will not matter. Now, we have only the second term which can be solved analytically by observing that for any square matrix $A$ with rank less than 1 or, more formally, $A\in\mathbb{R}^{D\times D}, \rho(A)<1$, we have $(I-A)^{-1}=I+A+A^2+\cdots$.
\begin{align}
    n_U = \left(I-\Rho^e_{UU}\right)^{-1}\Rho^e_{UK}\cdot n_K
\end{align}

Note that the inverse exist $(I-\Rho^e_{UU})^{-1}$ if every connected component has a label. As a consequence of doing the exponential operation, every node in the graph is connected to every other node and there will be no zero transition probabilities. Therefore, any principal submatrix of $\Rho^e$ will implicitly satisfy this condition too.

\begin{algorithm}
  \DontPrintSemicolon
  \SetKwProg{Fn}{Function}{:}{}
  \SetKwFor{uWhile}{While}{do}{}
  \SetKwRepeat{Do}{do}{while}
  \SetKwFor{ForPar}{For all}{do in parallel}{}
  \SetKwFunction{fitGNN}{fit-GNN}
  \SetKwFunction{predictGNN}{predict-attributes-GNN}
  \SetKwFunction{propagate}{KnowProp-GNN}
    \Fn{\fitGNN{$\textbf{n}_E, C$}}{
        $\{K, U\} \gets C$, {\small split into [known, unknown]}\;
        $\mathcal{G}_{nn}\gets$ MLP($\operatorname{Units}: \operatorname{In}$=$E, \operatorname{Out}$=$C$)\;
        $X\gets n_E[K]\in\mathbb{R}^{K\times E}$, $\operatorname{train-batch}$\;
        \small{**\text{Train via cross-validation}**}\;
        \Do{$\mathcal{L}_r > \epsilon$}{
            $K^P$ = $\mathcal{G}_{nn}(X)$ ,~~~{\small (Predict label distributions)}\; 
            $\mathcal{L}_r$ = $\norm{K - K^p}_2^2$\;
            $\mathcal{G}_{nn}\gets $ updated by backprop on $\mathcal{L}_r$\;
        }
     \KwRet $\mathcal{G}_{nn}$
    }\vspace{2mm}
    \Fn{\propagate{$\Rho, C$}}{
        Input: $\Rho\in\mathbb{R}^{D\times D}$, $C\in\operatorname{cat}^{D\times 1}$\;
        $\Rho^m_{i,j} = \frac{e^{\alpha\Rho_{i,j}}}{\max_{i,j}(e^{\alpha\Rho})}$~$(\operatorname{max-normalization})$\;
        \small{**Obtaining node embeddings of size $E$**}\;
        $\textbf{n}_E\gets\operatorname{weighted-node2vec}(\Rho^m)\in\mathbb{R}^{D\times E}$\;
        $\mathcal{G}_{nn}\gets$\fitGNN$(\textbf{n}_E, C)$\;
        $\{K, U\} \gets C$, {\small split into [known, unknown]}\;
        $U^p\gets\mathcal{G}_{nn}(\textbf{n}_E[U])$~~(predict unknown labels)\;
        \KwRet $U^p$
    }
\caption{GNN-based KP over CI graph}\label{algo:prop-algo-gnn}
\end{algorithm}

Alg.~\ref{algo:prop-algo} provides both the iterative and the analytical version of the algorithm for attribute propagation over the conditional independence graph. The iterative procedure can be helpful to get approximate solutions for very large graphs (where the number of nodes D is high) as the time complexity is roughly linearly proportional to the number of iterations needed for convergence. On the other hand, the analytical solution is dominated by the matrix inverse term which is approximately $O(D^3)$. One can work out the preferred method based on the input graph size, available compute and runtime constraints.



\subsection{Deep learning approach}

We can alternatively use Graph Neural Networks to increase the expressiveness and more fine grain control over the information that we propagate over the graph. The exponential scaled partial correlation matrix can act as the transition probability matrix in the weighted node2vec~\cite{grover2016node2vec}. The explore-vs-exploit strategy of the node2vec which explores the neighbor embeddings to be included for the node update will take into account the partial correlations that influence relationship between different input features. Alg.~\ref{algo:prop-algo-gnn} describes the procedure followed to use GNNs for knowledge propagation. The node2vec module provides the embeddings $E$ of the individual nodes that captures the graph structure provided by the probability transition matrix $\Rho$. After obtaining the embeddings, the \texttt{fit-GNN} function learns a neural network over the known features to predict their categories $C$. After training, we use this trained NN to predict the labels for the unknown features.

\section{Experiments}


\subsection{Datasets}

\textbf{Cora.} We use the Cora dataset~\cite{sen:aimag08}, which consists of 2708 scientific publications classified into one of seven classes.  Each publication in the dataset is described by a 0/1-valued word vector indicating the absence/presence of the corresponding word from the dictionary. The dictionary consists of 1433 unique words.  We generate graphs and subsets of Cora, with 300 nodes each for the experiments.

\textbf{PubMed Diabetes.} The dataset~\cite{sen:aimag08} consists of 19717 scientific publications from PubMed database pertaining to diabetes classified into one of three classes. Each publication is described by a TF/IDF weighted word vector from a dictionary which consists of 500 unique words.  We generate graphs and subsets of 300 nodes for experiments.

\subsection{Algorithms used in experiments}

In the experiments, we compared several variants of the knowledge propagation algorithms proposed in this paper.  The results are shown in Table~\ref{tab:inference_results}. 
The algorithms used are:
\begin{itemize}
\item node2vec is the algorithm described in Alg.~\ref{algo:prop-algo-gnn}, which follows~\cite{grover2016node2vec}'s procedure with partial correlation matrix as input.  We used settings $p=1,q=2$. These settings were narrowed down based on the performance on validation set. The embedding size was chosen to be $64$. We tried different embedding sizes: $\{32, 64, 128, 256\}$. Logistic regression classifier was trained over the embeddings to predict the final classes. Different classifiers were tried including MLP, which after substantial fine-tuning performed equivalent to the LR model.
\item iterative-exp is the iterative version of our algorithm using the transition probability matrix created using entry-wise exponential of the partial correlation matrix.
\item iterative-pos is the iterative version of our algorithm using the transition probability matrix created using positive correlation entries only as described in Eq.~\ref{eq:pos}, with KL regularization term. The update terms using $\Rho^-$ are omitted from the update in Eq.~\ref{eq:iterative-update}.
\item iterative-posneg is the iterative version of our algorithm using the transition probability matrices created using both positive and negative entries in the partial correlation matrix and update from Eq.~\ref{eq:iterative-update}.
\item analytical-exp is the analytical solution, refer Sec.~\ref{sec:analytical-proof}.
\end{itemize}

\subsection{Prediction accuracy}

All tests were performed on subsets of size 300.  We optimized hyperparameters (choice of regularization, value of $\alpha$) on graphs built on validation data and evaluated on graphs built on test data. 
All our algorithms outperform word2vec, with the exception of iterative-pos (the one ignoring negative correlations) on Cora with 20\% and 40\% missing values.  Analytical solution performs best on Cora, narrowly beating iterative solution using the same transition probability matrix. Note that these two methods, analytical and iterative method (without regularization) using the same $\Rho^e$ transition matrix should in principle result in the same attribute assignment. In practice, however, the iterative method is never run to the limit and is expected to give slightly inferior results. Iterative-posneg outperforms all other algorithms on the PubMed dataset.

\begin{table*}[]
\centering
\caption{\small Comparison of predictive accuracy for attribute propagation algorithms. 
 Hyperparameters optimized on validation set, results reported on the test set.
 Averaged over 50 runs with the means and standard deviations being reported.} 
\label{tab:inference_results1}
\resizebox{0.99\textwidth}{!}{
\begin{tabular}{|c|c|c|c|c|c|c|}
\hline
Methods & \multicolumn{3}{c|}{Cora (7 classes)} & \multicolumn{3}{c|}{PubMed Diabetes (3 classes)}  \\ 
 \hline
& \multicolumn{3}{c|}{missing attributes} & \multicolumn{3}{c|}{missing attributes} \\ \hline
& 20\% & 40\% & 60\% & 20\% & 40\% & 60\% \\ \hline
node2vec & $0.4233\pm0.0596$ & $0.3958\pm0.0352$ & $0.3655\pm0.0297$ & $0.5740\pm0.0657$ & $0.5429\pm0.0432$ & $0.5147\pm0.0336$  \\ \hline
iterative-exp & $0.5527\pm0.0542$ & $0.5183\pm0.0418$ & $0.4664\pm0.0259$ & $0.5853\pm0.0551$  & $0.5912\pm0.0425$ & $0.5706\pm0.0363$ \\ \hline
iterative-pos & $0.3910\pm0.0724$	& $0.3710\pm0.0427$ & $0.3712\pm0.0416$ & $0.6223\pm0.0696$ & $0.5987\pm0.0677$  & $0.5769\pm0.0553$  \\ \hline
iterative-posneg & $0.5210\pm0.0697$ & $0.4793\pm0.0572$ &$0.4350\pm0.0497$ & {\boldmath $0.7227\pm0.0476$} & { \boldmath $0.6770\pm0.0392$} &	{\boldmath $0.6182\pm0.0240$}  \\ \hline
analytical-exp & {\boldmath $0.5683\pm0.0592$} & {\boldmath $0.5275\pm0.0428$ }& {\boldmath $0.4667\pm0.0308$} & $0.5859\pm0.0637$  & $0.5658\pm0.0394$ & $0.5398\pm0.0333$ \\ \hline
\end{tabular}}
\label{tab:inference_results}
\end{table*}

\subsection{Additional experiments}

\begin{figure}
\centering 
\includegraphics[width=0.45\textwidth]{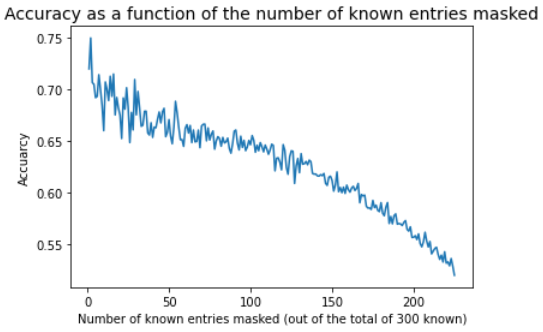}
\caption{\textbf{\texttt{iterative-posneg} on CORA}:
The graph shows the accuracy on the Cora validation dataset as a function of the number of missing attributes. The range varies from 1 node with a missing attribute to 225 nodes (75\% of all nodes). The method used is the iterative solution consisting of both positive and negative transition matrices with KL-divergence as the regularization mechanism. Results are averaged over 50 runs and the mean accuracy is reported.}
\label{fig:iterative-KL-posneg}
\end{figure}

Fig.~\ref{fig:iterative-KL-posneg} shows algorithm accuracy as a function of the number of known entities masked, from 1 to 75\% of all nodes.  The accuracy declines, as expected, from 0.75 to below 0.55, steeply from 1 to 50, more gradually from 50 to 120 and again more steeply after that.  The accuracy fluctuates most for low number of masked entities. 

\begin{figure}
\centering 
\includegraphics[width=0.45\textwidth]{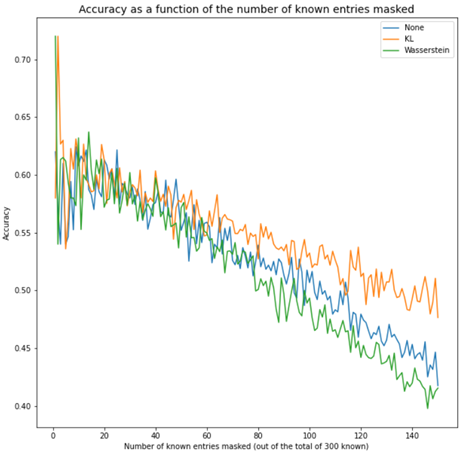}
\caption{\textbf{\texttt{iterative-pos} on CORA with different regularization options}: We restrict the associated CI graph recovered for the CORA data to only the positive partial correlations. The plot shows the accuracy on the Cora validation dataset as a function of the number of missing attributes ranging from 1 node with a missing attribute to 150 nodes (50\% of all nodes). The three curves show the results of applying three regularization options: KL-divergence, Wasserstein's distance and no regularization mechanism. Results are averaged over 50 runs and the mean accuracy is reported.}
\label{fig:iterative-KL-pos}
\end{figure}

Fig.~\ref{fig:iterative-KL-pos} shows the difference the regularization term makes, showing accuracy as a function of known entities masked for iterative-pos algorithm, with KL-divergence term, with Wasserstein's distance term and no regularization.  KL-divergence regularization works best on this dataset.  The difference becomes more pronounced for a larger number of entities masked.

\begin{figure}
\centering 
\includegraphics[width=0.45\textwidth]{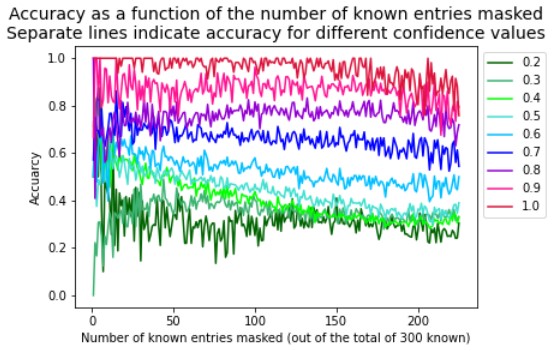}
\caption{\textbf{\texttt{iterative-pos} on CORA with different confidence bands}:
The accuracy is calculated on the Cora validation dataset as a function of the number of missing attributes, ranging from 1 node with a missing attribute to 225 nodes (75\% of all nodes). The method used is the iterative solution with positive transition matrix and KL regularization mechanism. Results averaged over 50 runs.}
\label{fig:iterative-KL-pos-confidence}
\end{figure}

The last figure, Fig.~\ref{fig:iterative-KL-pos-confidence} shows the result of the experiment designed to test the feasibility of reporting attribute value assignments for nodes we can predict with high confidence.  Recall that the output of the procedure is a distribution over possible attribute values.  If we report the result for entries where the maximum value in the distribution exceeds the given threshold, we can achieve much higher accuracy, which may be the desired outcome in some domains where accuracy is more important than coverage.

\section{Conclusions}

Conditional Independence graphs are particularly useful for domain exploration tasks, with several CI graph recovery methods and applications being proposed recently~\cite{shrivastava2022methods}. Much work has been done to make the CI graph recovery algorithms fast and scalable. We are hoping to extend the usefulness of these models by providing algorithms to reason about the domain based on the recovered graphs.  In this work we proposed a set of methods, iterative and analytical, to assign attribute values to nodes where these values are unknown.  The experiments demonstrate that our algorithms improve upon the state-of-the-art node2vec method, even when it is informed by the partial correlation matrix.   

We would like to direct attention of our readers to the fact that we were able to find better regularization terms to improve the performance of the KP algorithm which we demonstrated empirically. In our $\operatorname{iterative-posneg}$ procedure that is used in Alg.~\ref{algo:prop-algo}, we discovered that splitting the KL-divergence penalty over the positive and negative correlation gives considerable improvements in some cases, refer to the results of the PubMed Diabetes experiment (Table~\ref{tab:inference_results}). We are currently exploring possibilities of designing a more generic regularization term which can achieve better generalization results over a variety of domains.




\bibliography{citations,bibfile}

\clearpage

\end{document}